\documentclass[12pt]{article}

\usepackage{scicite}

\usepackage{times}

\newenvironment{sciabstract}{%
\begin{quote} \bf}
{\end{quote}}

\newcounter{lastnote}

\usepackage{xcolor}
\definecolor{darkgreen}{RGB}{30,150,30}

\usepackage{ragged2e}
\usepackage{tcolorbox}
\usepackage{float}

\usepackage{graphicx}
\usepackage{grffile} 

\usepackage{authblk}
\usepackage{relsize}

\usepackage{subcaption}

\usepackage{caption}

\usepackage[
    hidelinks,
    colorlinks=true, 
    linkcolor=black, 
    urlcolor=darkgray, 
    pdfauthor={author name}, 
    pdftitle={Document Title}, 
    pdfsubject={Subject}, 
    pdfkeywords={keywords}
]{hyperref}

\usepackage{siunitx}

\usepackage{booktabs}
\usepackage{array}

\title{Enabling robots to follow abstract instructions and complete complex dynamic tasks}

\author
{Ruaridh Mon-Williams,$^{1,2,3}$ Gen Li$^{1\ast}$ Ran Long$^{1\ast}$ Wenqian Du$^{1, 4 \ast}$, Chris Lucas$^{1}$\\
\normalsize{$^{1}$University of Edinburgh, $^{2}$Massachusetts Institute of Technology, $^{3}$Princeton University, $^{4}$Alan Turing Institute}
}

\date{}

\usepackage{graphicx}
\begin{document} 


\baselineskip24pt


\maketitle


\begin{sciabstract}
Completing complex tasks in unpredictable settings like home kitchens challenges robotic systems. These challenges include interpreting high-level human commands, such as “make me a hot beverage” and performing actions like pouring a precise amount of water into a moving mug. To address these challenges, we present a novel framework that combines Large Language Models (LLMs), a curated Knowledge Base, and Integrated Force and Visual Feed- back (IFVF). Our approach interprets abstract instructions, performs long-horizon tasks, and handles various uncertainties. It utilises GPT-4 to analyse the user’s query and surroundings, then generates code that accesses a curated database of functions during execution. It translates abstract instructions into actionable steps. Each step involves generating custom code by employing retrieval-augmented generalisation to pull IFVF-relevant examples from the Knowledge Base. IFVF allows the robot to respond to noise and disturbances during execution. We use coffee making and plate decoration to demonstrate our approach, including components ranging from pouring to drawer opening, each benefiting from distinct feedback types and methods. This novel advancement marks significant progress toward a scalable, efficient robotic framework for completing complex tasks in uncertain environments. Our findings are illustrated in an \href{https://youtu.be/WiLEw9Zu2MA}{accompanying video} and supported by an open-source GitHub repository (released upon paper acceptance).

\end{sciabstract}


\section*{Introduction}

Domestic settings present a sharp contrast to the controlled environments typical in industrial automation settings. The layout of homes is constantly changing, inhabited by individuals who frequently alter their surroundings. These ever-evolving conditions pose significant challenges for robotic systems, which need to continuously adapt to successfully interact with objects in such settings \cite{billard_trends_2019, cui_toward_2021, arents_smart_2022}. Robots in these environments are required to interpret high-level human instructions, manage tasks over extended periods (long-time horizons), and integrate force and visual feedback (IFVF) as a strategy to mitigate uncertainties, including environmental changes caused by human activity or sensor noise \cite{arents_smart_2022, billard_trends_2019}. We propose a novel methodology to address these challenges.\\

Consider a scenario where someone returns home feeling fatigued and wants a refreshing beverage. A robot with a sophisticated manipulation system is situated in the homeowner’s kitchen and is informed about their tiredness and instructed to prepare a drink. The robot decides that a reinvigorating cup of coffee is just what the human needs. This task, seemingly straightforward, encompasses a series of challenges and tests the limits of current robotic capabilities \cite{cui_toward_2021, arents_smart_2022, billard_trends_2019, yang_grand_2018, buchanan_online_2024}. First, the robot must interpret the command it receives and analyse its surroundings. Next, it may need to search the environment to locate a mug. This could involve opening drawers with unspecified opening mechanisms. Then, the robot must measure and mix the precise ratio of water to coffee. This requires fine-grained force control, and adaptation to uncertainty if, for example, the human moves the location of the mug unexpectedly \cite{billard_trends_2019, nikolaidis_efficient_2015}. This scenario is a canonical example of the multifaceted nature of complex tasks in dynamic environments. Robotic systems have traditionally struggled with these tasks because they have relied on pre-programmed responses and lack the flexibility to adapt seamlessly to perturbations \cite{saveriano_dynamic_2023, kober_movement_2010}.

\begin{figure}[H]
\centering
\href{https://youtu.be/WiLEw9Zu2MA}{\includegraphics[width=\linewidth]{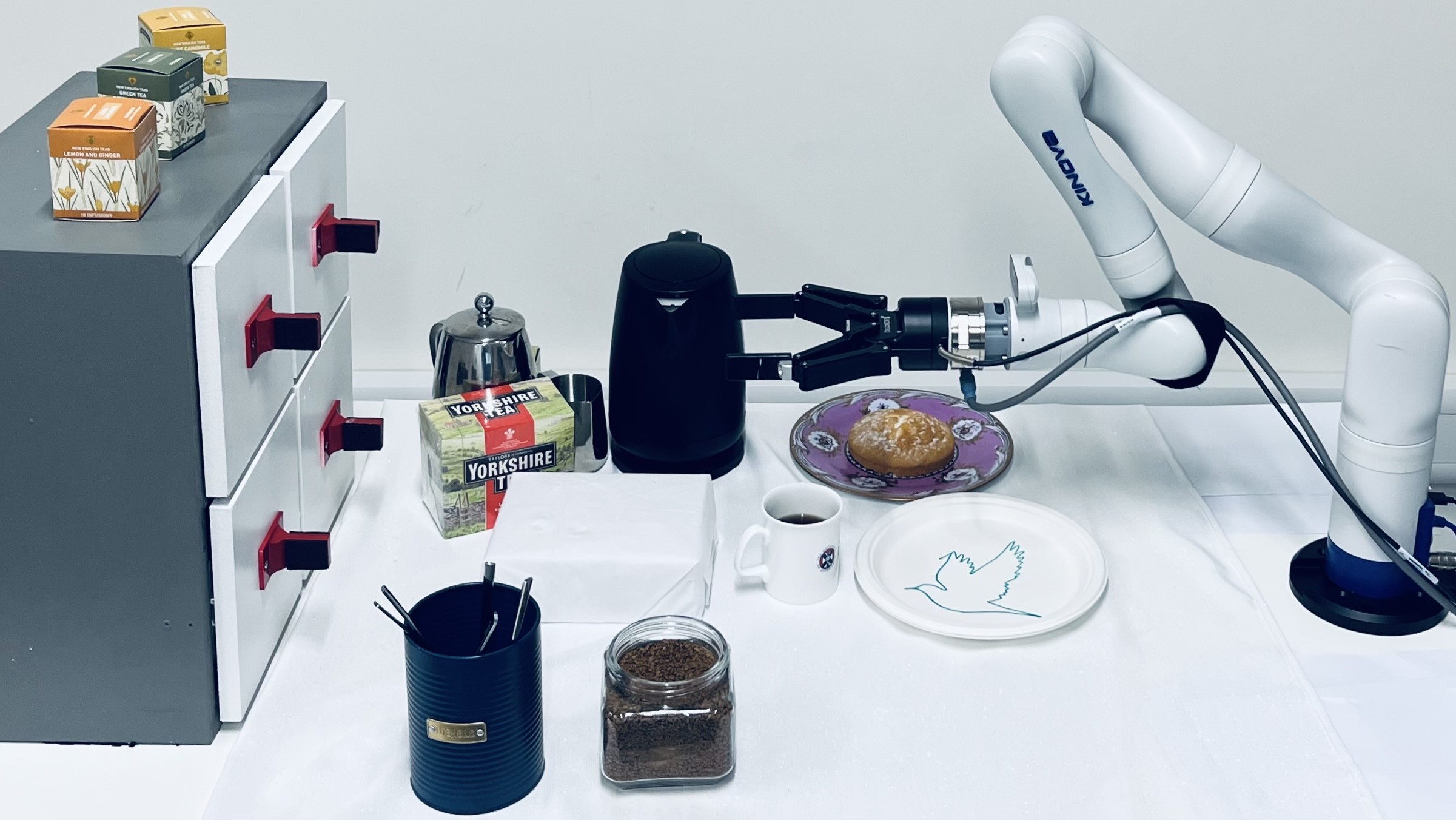}}
\caption{\textbf{Coffee and plate decoration video}. Kinova Gen3 Robot prepares coffee and decorates a plate. Click image for video demonstration.}
\label{fig:robot making coffee}
\end{figure}

Large Language Models (LLMs) provide a potential solution to these challenges \cite{ahn_as_2022, driess_palm-e_2023, peng_preference-conditioned_2024}. LLMs offer a way to process complex instructions and adapt actions accordingly because of their advanced contextual understanding and generalisation abilities \cite{huang_language_2022, huang_towards_2023}. A substantial body of work has explored the application of pre-trained language models for embodied agents \cite{huang_language_2022, ahn_as_2022, huang_voxposer_2023, zeng_socratic_2023, cui_no_2023, huang_inner_2023, nair_r3m_2022, singh_progprompt_2022, song_llm-planner_2023, vemprala_chatgpt_2023, driess_palm-e_2023, ding_task_2023}. This work has achieved impressive results in enabling robotic manipulation systems to comprehend context and apply their robotic skills across a wide range of tasks, predominantly those with short time-horizons, like pick-and-place operations, across various scenarios.\\

However, despite achieving proficient results in discrete tasks, there remains a gap in these systems’ ability to handle more complex, long-horizon tasks (e.g., coffee preparation) in dynamic, human-centric environments \cite{arents_smart_2022}. The robot must adapt its strategy depending on its surroundings, the task, and the involvement of humans. These factors could influence the type of coffee prepared (e.g., instant or filtered) and the role the robot assumes. Current LLM-driven approaches often rely on detailed prompts to correctly guide robot actions. However, the reliance on these lengthy prompts and inefficient feedback mechanisms means systems often struggle with complex, long-horizon tasks that require a diverse set of skills across various scenarios \cite{ahn_as_2022, raiaan_review_2024}. Furthermore, recent approaches often neglect the integration of force and visual feedback \cite{huang_voxposer_2023, raiaan_review_2024}. This integration is crucial in scenarios such as pouring water into a moving cup, where visual information is necessary to track the cup and force feedback is needed for pouring the desired amount of water despite visual occlusion \cite{rozo_force-based_2013, zhang_explainable_2022, huang_robot_2021}. Thus, there is a need for an innovative approach in robot manipulation that tackles issues including interpreting abstract instructions, managing long time horizons, and integrating visual and force feedback (IFVF) to effectively execute actions in the face of noise and other uncertainties.\\

We have developed a novel framework that combines the cognitive capabilities of LLMs with the dexterity of robotic systems to address these challenges. Specifically, the approach uses code as dynamic policies \cite{liang_code_2023} that can facilitate adaptable robotic actions. LLMs’ capabilities are leveraged using Retrieval-Augmented Generation (RAG) \cite{lewis_retrieval-augmented_2020} to dynamically select and adapt the most suitable policy from a database or generate its own code based on relevant examples. In contrast to existing pure LLM-driven methods \cite{huang_voxposer_2023, raiaan_review_2024,brohan_rt-2_2023}, we integrate force and vision into the framework, allowing the system to adapt to a variety of complex tasks in dynamic settings. This approach equips the robotic system with the capacity for high-level contextual understanding \cite{brohan_rt-2_2023} and the proficiency to execute complex tasks with real time feedback, ensuring accuracy and precision. The approach ensures that each action is aligned with the specific demands of the task and the environmental conditions. To demonstrate the framework’s capabilities, a 7-degree-of-freedom Kinova robotic arm was employed to execute complex, force intensive tasks in uncertain environments, leveraging integrated force and vision feedback. The overall system diagram is presented below.

\begin{figure}[H]
    \centering
    \includegraphics[width=1\textwidth]{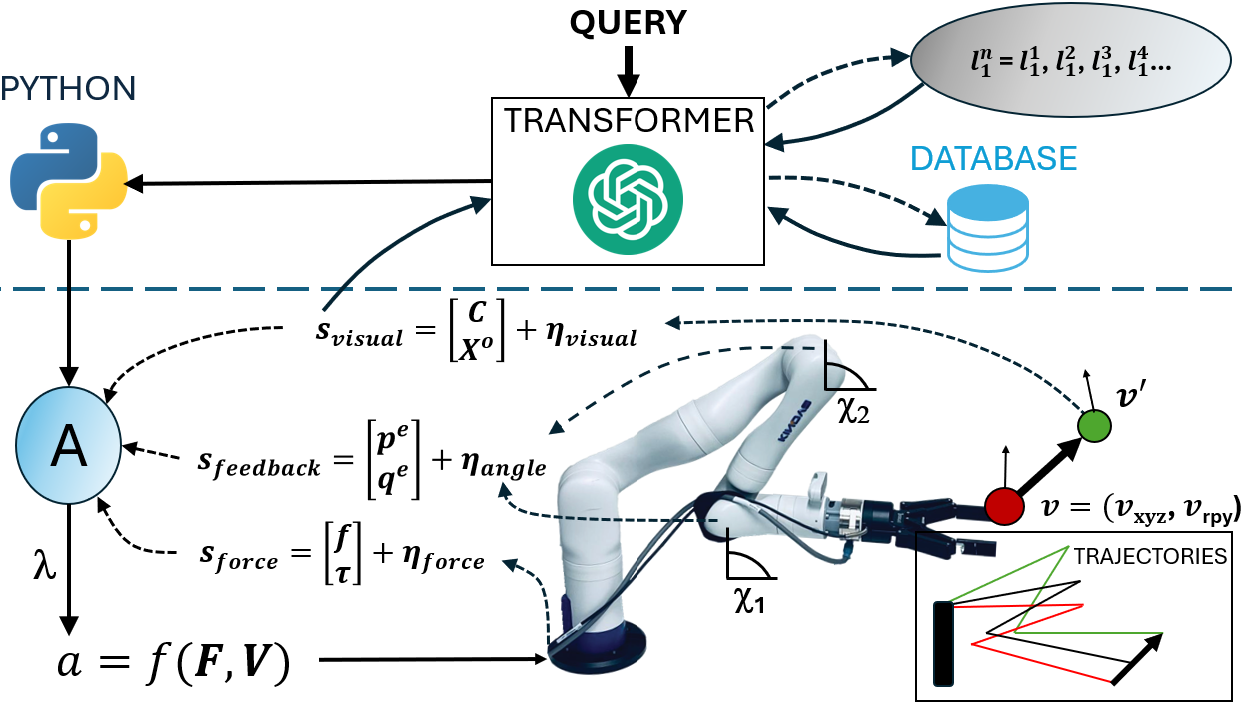}
    \caption{\textbf{Schematic of the system framework.} The schematic illustrates the system framework, showing the high-level (above the blue dashed horizontal line) and low-level (below the blue dashed horizontal line) system architecture. User queries are fed into a transformer via voice recognition software. The transformer (GPT-4) takes this input and integrates it with: \textbf{(i)} an image ($C$) of the environment (via an azure Kinect depth camera); \textbf{(ii)} knowledge base of code examples, including various functions stored in a database. The transformer decomposes the higher order abstracted task into actionable high-level subtasks, retrieves relevant code examples from the knowledge base, adapts them and writes python code tailored to these tasks. This code is then sent to the robot controller \textbf{(A)}. The controller processes the code and sends control signals ($\lambda$) to the robot. The actions \textbf{(a)} are controlled with force (\textbf{F}) and vision (\textbf{V}) feedback. The model uses vision to identify the properties of different objects (e.g., pose, ($X$), of a coffee cup), so it can grasp objects accurately. The robot uses force ($f$) and torque ($\tau$) feedback (available via an ATI force transducer) to manipulate objects skillfully (e.g., determine how much water to pour). Feedback is necessary due to noise within the vision signal (\textbf{$\eta_{vision}$}), the robot joint angles (\textbf{$\eta_{angle}$}), and the force transducer signal (\textbf{$\eta_{force}$}. The feedback updates the motion in the Robot Operating System (ROS) to achieve the desired goal through velocity commands of both linear ($v_{xyz}$) and angular ($v_{rpy}$) velocities. These commands generate trajectories based on appropriate forces and spatiotemporal patterns to achieve the sub-goals. The use of feedback loops, including 40 Hz updates of the end-effector position ($p$) and orientation ($q$), allow the robot to respond to disturbance (e.g., the robot tracking a cup to determine its new position after it is moved by the user).}
    \label{fig:enter-label}
\end{figure}

In conclusion, our framework integrates language processing, force and vision to enable robots to adapt to complex tasks. It is novel in its ability to combine the following features: (i) interpret high-level human commands; (ii) complete long-horizon tasks; and (iii) utilise IFVF to manage noise and disturbances in changing environments. This framework represents a significant step forward in robotics. It integrates force and vision feedback to deal with uncertainty, offers scalability, and allows the strengths of various methods to be combined. Methods can be incorporated from reinforcement learning, imitation learning, and flexible motion primitives to improve adaptability for diverse and dynamic scenarios.




\subsection*{Related Work}
Robot manipulation, a fundamental research problem, has witnessed significant advancements over the past decades. Traditional approaches often relied on predefined motion primitives and have limited ability to adapt to novel environments \cite{saveriano_dynamic_2023}. Motion primitives remain useful but major progress has been made through powerful alternatives including reinforcement learning and imitation learning \cite{shridhar_cliport_2022, shridhar_perceiver-actor_2022, mees_what_2022, mees_grounding_2023, mon-williams_behavioural_2023, shao_concept2robot_2020, belkhale_data_2024}. These methods have demonstrated the effectiveness of interaction and demonstration in teaching robots to perform complex tasks. However, these approaches, although promising \cite{huang_voxposer_2023}, often struggle to handle a wide variety of scenarios and novel tasks \cite{ahn_as_2022} and require extensive data collection and training \cite{khazatsky_droid_2024}. This limitation has restricted robots’ efficiency in dynamic real-world scenarios, where there is a need for adaptation to unforeseen changes.\\

Reinforcement learning (RL) methods often struggle in adapting to diverse scenarios due to their reliance on specific, often narrow, training environments. This limits their ability to dynamically adapt to a wide variety of contexts. RL also requires extensive training which can be costly unless simulators are used. Moreover, the simulators used for RL system training frequently encounter instabilities and often struggle to efficiently and accurately replicate the subtleties of real-world physics in dynamic scenarios \cite{acosta_validating_2022, alomar_causalsim_2023, choi_use_2021}, such as the manipulation of fluids or granular materials \cite{rozo_force-based_2013, zhang_explainable_2022, huang_robot_2021}. While RL has proven effective for simpler tasks, such as box pushing \cite{del_aguila_ferrandis_nonprehensile_2023}, its application over a wide range of complex and varied scenarios is less successful. Techniques to address these issues include domain randomisation, meta-learning and extensive training over a wide range of environments \cite{kirk_survey_2023, dai_analysing_2022}. However, these methods still struggle to adapt to a wide variety of tasks beyond their original training environments, highlighting a persistent challenge in creating broadly applicable and adaptable RL systems.\\

Imitation learning (IL), despite offering a practical solution by which a robot can perform a wide range of manipulation tasks \cite{zhang_explainable_2022, hussein_imitation_2017, di_palo_dinobot_2024, shridhar_cliport_2022, shridhar_perceiver-actor_2022, mees_what_2022, mees_grounding_2023, shao_concept2robot_2020}, faces challenges when adapting to new contexts. IL is adept at capturing subtle behaviours and can produce ‘human-like’ motions, but it struggles with distribution shift, requiring extensive diverse data to be truly effective beyond the specific data collection conditions \cite{chang_mitigating_2021}. Moreover, many IL models require extensive additional training data to account for various unexpected changes in the environment, limiting robots to primarily replicate the learnt tasks without dynamically adapting to uncertainties.\\

The recent emergence of Large Language Models (LLMs) has offered opportunities to overcome these traditional limitations in robotic capabilities \cite{brohan_rt-2_2023, zhang_large_2023, ahn_as_2022, kwon_toward_2024, hong_learning_2024} ADDCITATIONS. LLMs introduce versatility and decision-making capabilities that were previously unattainable, enabling robots to manipulate objects in complex environments with enhanced adaptability \cite{ahn_as_2022}. A significant body of recent research has used LLMs for short-horizon tasks \cite{huang_voxposer_2023, huang_language_2022, brohan_rt-2_2023} ADDCITATIONS. For instance, VoxPoser utilises LLMs to perform a variety of everyday manipulation tasks and provides an extensive literature survey on related work \cite{huang_voxposer_2023}. Similarly, the Robotics Transformer (RT-2) leverages large-scale web data and robotic learning data, enabling it to perform tasks beyond its training scope, demonstrating remarkable adaptability \cite{brohan_rt-2_2023}. Moreover, the Hierarchical diffusion policy (HDP) introduces a model structure to generate context-aware motion trajectories, which enhances task-specific motions from high-level LLM decision inputs \cite{ma_hierarchical_2024}. However, challenges remain in effectively integrating LLMs into robotic manipulation. These include complex prompting requirements, a lack of real-time interacting feedback, a dearth of LLM-driven manipulation work exploiting the use of force in robotic manipulation, and inefficient pipelines that reduce the seamless execution of tasks \cite{huang_voxposer_2023, zhang_large_2023}. These constraints limit the extent to which LLM-driven robotic manipulation can achieve high degrees of precision and adaptability. Thus, combining the contextual understanding of LLMs with the strength of other motion policies (e.g., RL, imitation learning, task specific functions etc.) provides an exciting possible solution to enhance robotic capabilities.\\

One powerful possibility made possible through LLMs is their ability to access and utilise extensive ‘knowledge bases’. Leveraging knowledge bases to enhance LLMs’ potential for robotic applications has already shown substantial benefits \cite{vemprala_chatgpt_2023}. This method enables robots to access and retrieve relevant action examples and information, assisting precise response generation. Nonetheless, approaches have been limited by the size of the knowledge base, and hence its diversity, due to performance decreases when adding large knowledge bases directly into the LMM’s context window \cite{huang_voxposer_2023, vemprala_chatgpt_2023}. Retrieval-augmented generation (RAG) represents a significant advancement in overcoming this limitation and taking full advantage of LLMs \cite{lewis_retrieval-augmented_2020}. However, the application of RAG in robotics has not been widely explored despite its potential to continually update and refine robot knowledge with relevant and accurate examples (and increase the knowledge base size without impacting performance).\\

Our approach addresses the limitations of existing methods by synergistically combining language processing with visual and force feedback in a framework that can exploit the strengths of different methods. We use LLMs, augmented with feedback loops and retrieval augment generation (RAG), to write expressive code and facilitate complex manipulation tasks. The approach enables real-time adaptation to environmental changes and leverages a repository of precise solutions via RAG. This ensures accurate task execution and meets the requirements for a system with broad adaptability \cite{lewis_retrieval-augmented_2020}. Our approach distinguishes itself by integrating not only vision and language but also force feedback. This approach allows the robot to execute complex long-horizon tasks and adeptly manage uncertainties.

Our method requires known constraints to be encoded into the code examples or motion functions, but is designed to rapidly accommodate and scale to numerous uncertainties. This adaptability ensures our method can respond to real-time challenges, such as fluctuating ingredient quantities or scenarios such as opening unknown drawers – capabilities that other methods lack without extensive additional training \cite{raiaan_review_2024,brohan_rt-2_2023,zeng_socratic_2023, cui_no_2023}.\\

Our example of coffee making and plate decorating only represents a small subset of the types of complex tasks that a sophisticated robot might be required to undertake. Nevertheless, this approach is conducive to being scaled up so it incorporates a wide range of possible long horizon tasks. It is practical to create a database of feedback loops or learning from demonstration examples that can be integrated together to allow for a wide variety of complex robotic manipulations.

\section*{Results}
\subsection*{Experimental Setup}
We employed an Azure Kinect DK Depth Camera, which was set to a resolution of 640x576 px with a sample rate of 30 FPS for depth sensing. Calibration was achieved using a 14cm AprilTag, allowing alignment between the camera and the robot’s base to an accuracy of less than $10^{-6}$. This setup enabled accurate object position detection within the scene.

For object interaction, we used an ATI multi-axis force and torque sensor. This provided six components of force and torque exerted by the robot’s end-effector during task execution. The sensor’s accuracy is within $\sim$2$\%$ of the full scale at a sampling rate of 100 Hz. For identifying the object pose, a 3D voxel was created. We used Grounded-Segment-Anything \cite{kirillov_segment_2023} for our language-to-vision module.

\subsection*{Task Description}
\begin{figure}[H]
    \centering
    \begin{subfigure}[t]{0.45\textwidth}
        \centering
        \includegraphics[width=\textwidth]{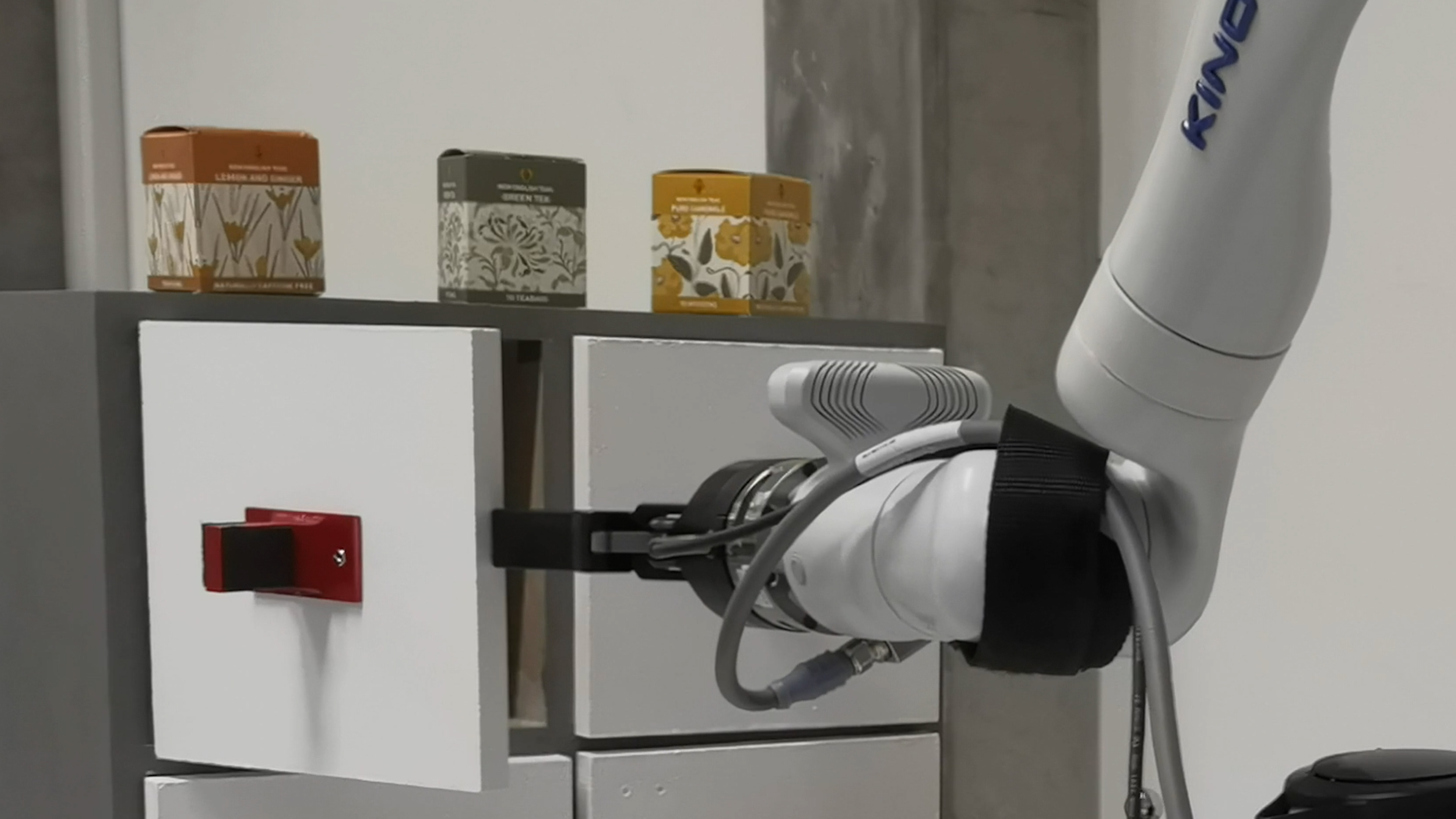}
        \caption{Door Opening}
        \label{fig:door_open}
    \end{subfigure}
    \hfill 
    \begin{subfigure}[t]{0.45\textwidth}
        \centering
        \includegraphics[width=\textwidth]{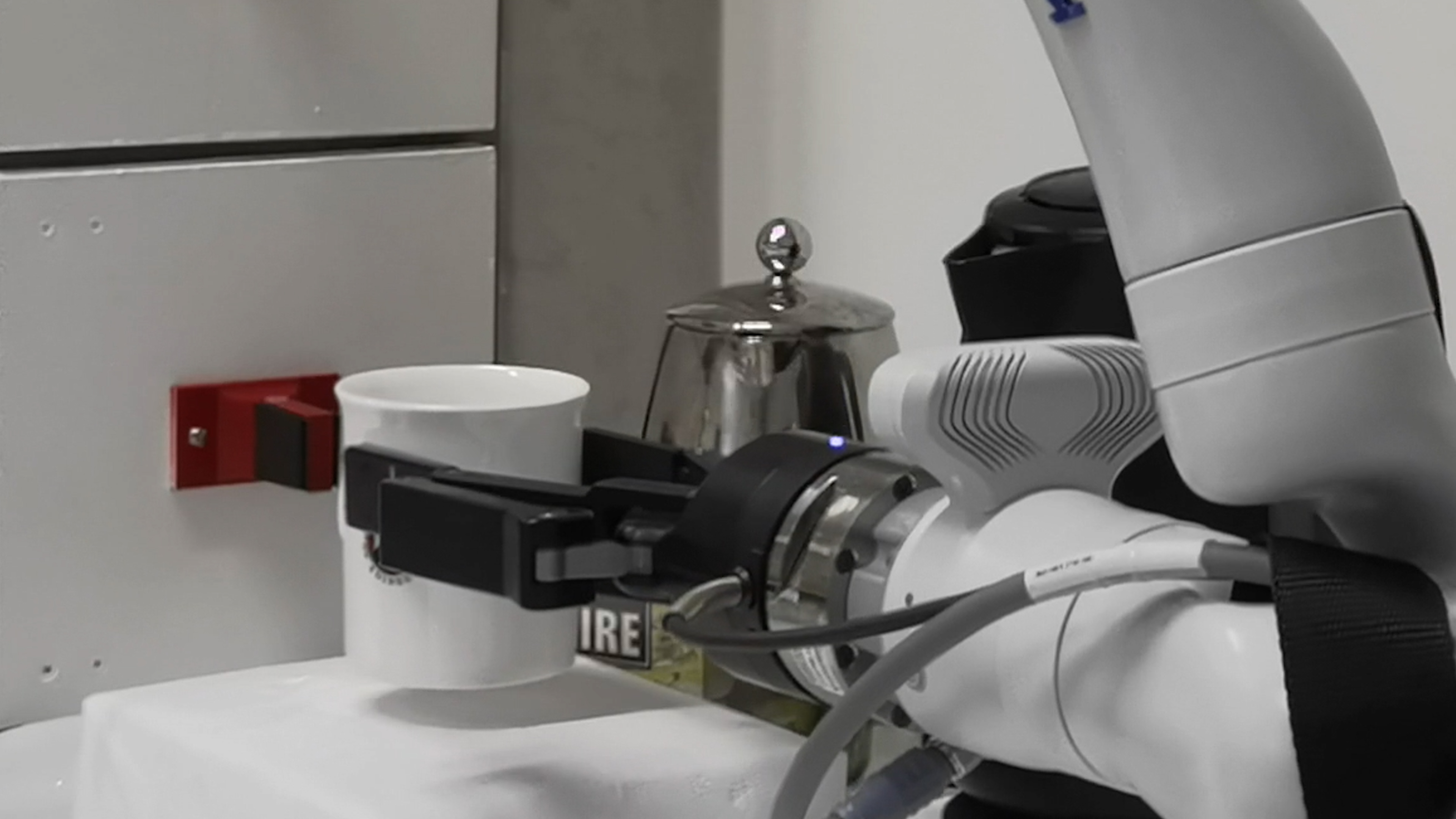}
        \caption{Pick and Place}
        \label{fig:pick_and_place}
    \end{subfigure}
    
    \begin{subfigure}[t]{0.45\textwidth}
        \centering
        \includegraphics[width=\textwidth]{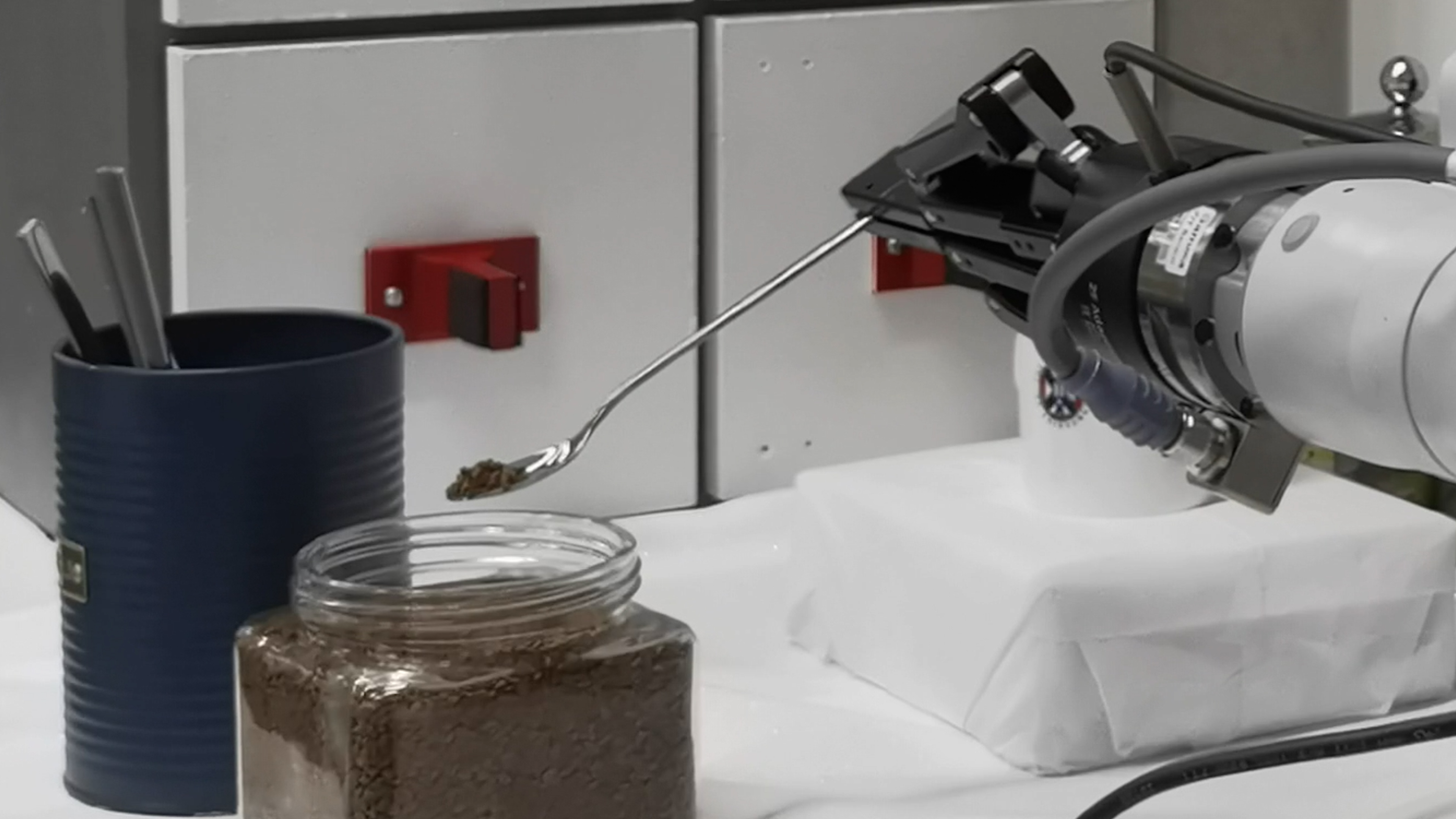}
        \caption{Scooping}
        \label{fig:scooping}
    \end{subfigure}
    \hfill 
    \begin{subfigure}[t]{0.45\textwidth}
        \centering
        \includegraphics[width=\textwidth]{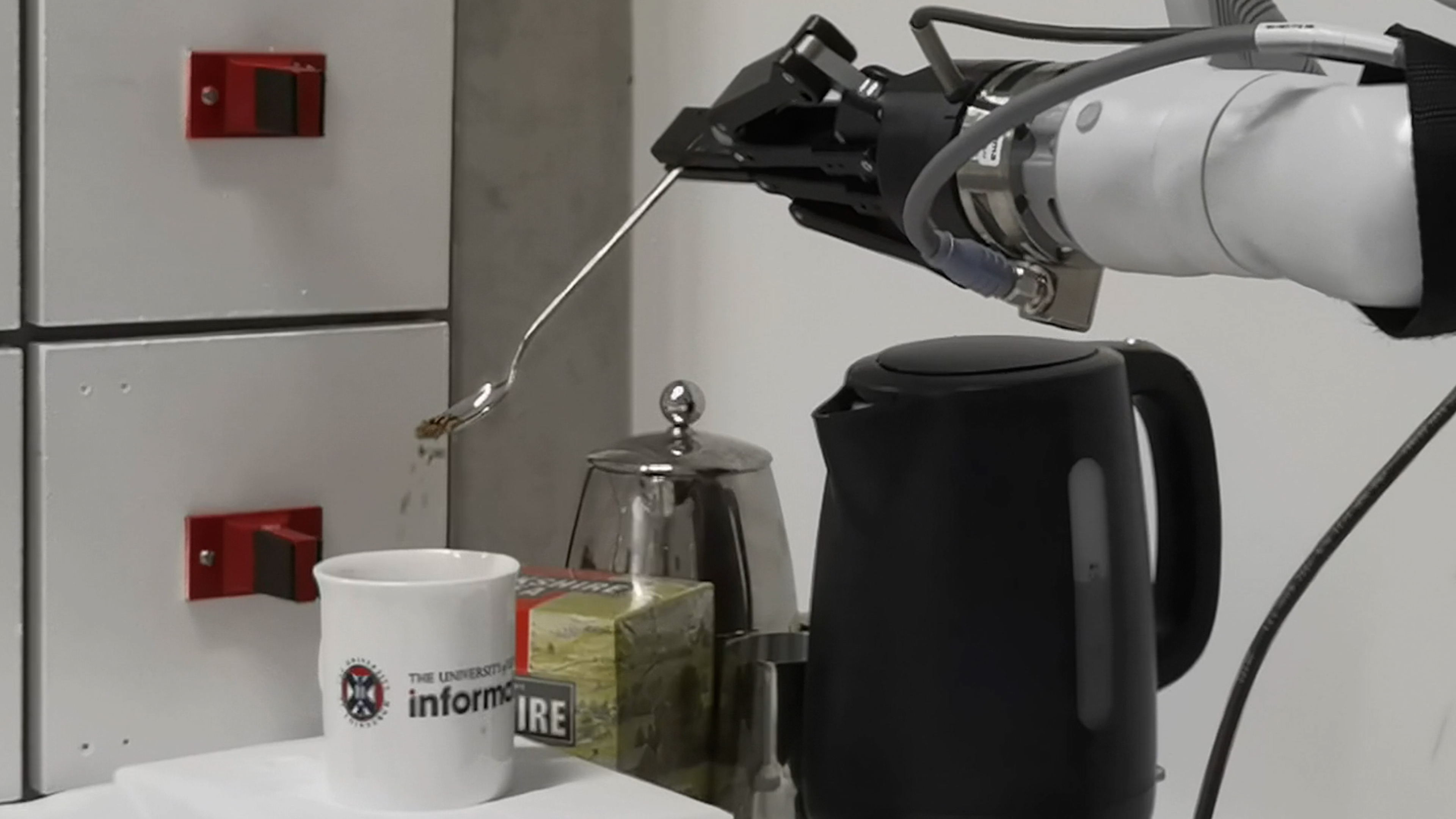}
        \caption{Emptying}
        \label{fig:emptying}
    \end{subfigure}

    \begin{subfigure}[t]{0.45\textwidth}
        \centering
        \includegraphics[width=\textwidth]{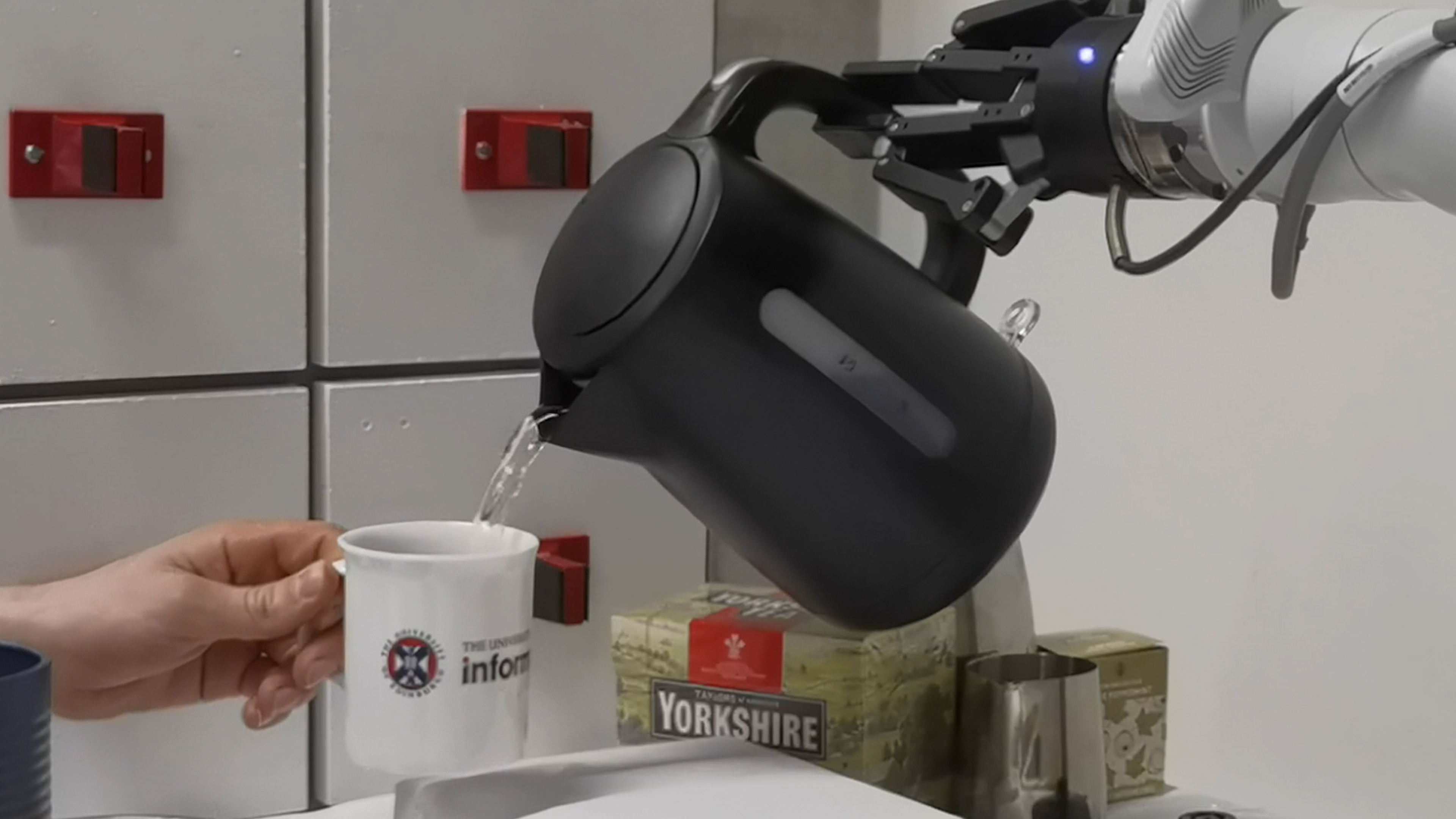}
        \caption{Pouring}
        \label{fig:pouring}
    \end{subfigure}
    \hfill 
    \begin{subfigure}[t]{0.45\textwidth}
        \centering
        \includegraphics[width=\textwidth]{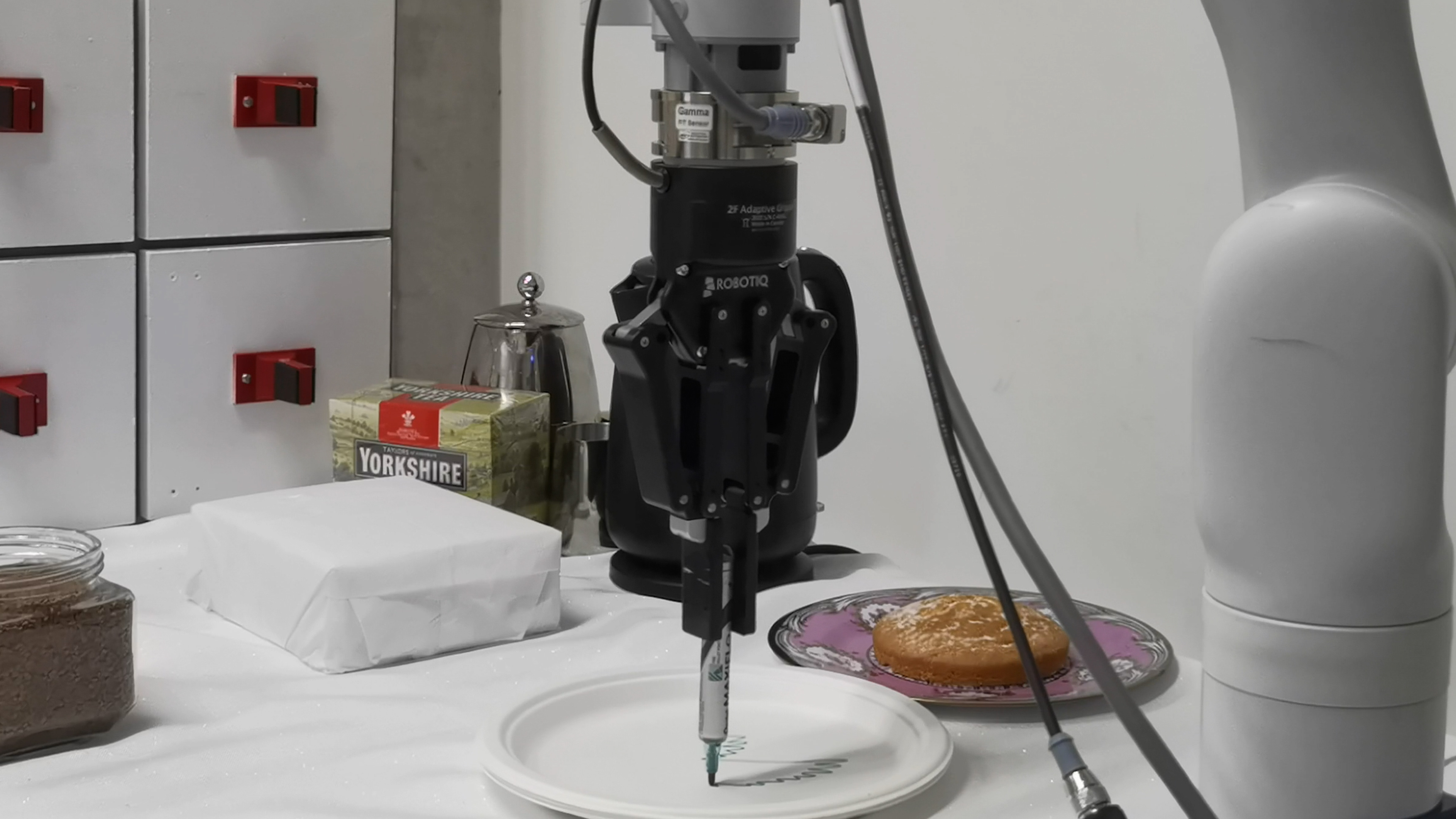}
        \caption{Drawing}
        \label{fig:drawing}
    \end{subfigure}

    \caption{Action shots of the Kinova Gen3 robot preparing coffee and decorating a plate.}
    \label{fig:robot_actions}
\end{figure}

The objective was to design the robot to capably perform any appropriate task specified by the user. To achieve this, we provided the robot with a comprehensive database of flexible examples. The database included a variety of flexible examples of specific motions, as illustrated by Figure \ref{fig:robot_actions}. The robot could replicate and adapt the motions to execute complex tasks requested by the user. The system was designed to enable the robot to dynamically adjust to environmental variables and uncertainties. This enhances the robot’s effectiveness in unpredictable conditions, thereby improving its flexibility and adaptability in real-world situations. Included in the database were examples of pouring liquids, scooping powders, opening doors with unknown mechanisms, picking up and placing down objects, drawing anything requested, conducting handovers, and moving in various directions, orientations, or relative to specified objects. This diverse functionality underscores the robot’s adaptability and ability to task a diverse range of practical challenges, such as making coffee and decorating a plate.

\subsection*{Language Integration for Abstract Reasoning}

To equip the robot with capabilities for abstract reasoning, we integrated GPT-4 \cite{openai_gpt-4_2024}, a language model that enables the robot to process user queries and environmental data to break down tasks into actionable steps. This involves generating code and executing actions with force and vision feedback, effectively providing the robot with a form of intelligence. We created a custom GPT-4 \cite{openai_gpt-4_2024, openai_custom_2023} and provided it with our database of flexible motion examples. This includes pouring, scooping, drawing, handovers, pick-and-place, and opening doors. Using retrieval-augmented generation (RAG), the robot could identify and extract relevant examples for the downstream task. The curated knowledge base, combine with RAG, allows the language model to access a vast selection of low and high-order functions for many situations, each with known uncertainties. This capability enables the robot to effectively handle numerous scenarios. The architecture is designed for flexibility, allowing for straightforward integration of additional functionalities such as reinforcement learning and learning from demonstration, enhancing the robot’s adaptability to perform additional tasks.

\subsection*{Zero-Shot Pose Detection}
The vision system generates a three-dimensional voxel representation. This contains the meshes of various objects. From these meshes, target poses are extracted at a frequency of $1/3$ Hz. The system can detect any object in principle; however, in practice, it does not always accurately identify each object. This is often due to confusion between objects with similar shapes or objects absent from the training dataset. Moreover, occlusion caused by the robot’s end-effector can result in inaccuracies in object detection and can lead to errors in the sensory data.

\begin{figure}[H]
\centering
\includegraphics[width=\linewidth]{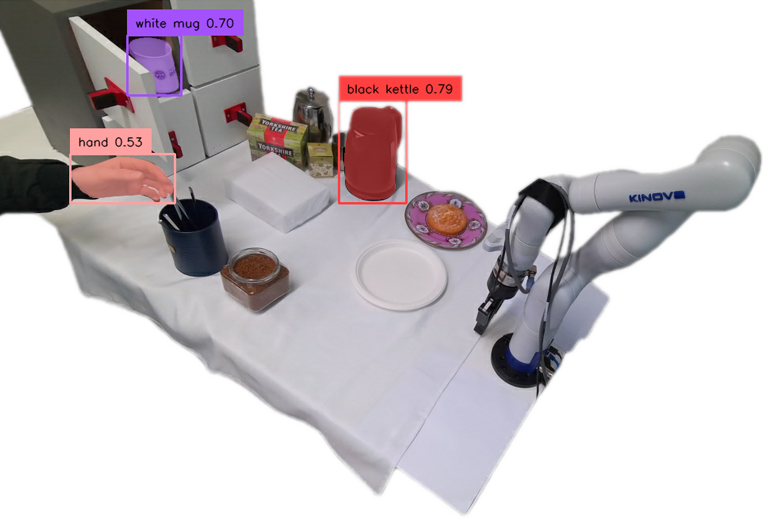}
\caption{\textbf{Vision detection module.} Illustration of the zero-shot vision detection module identifying a hand, white mug, and black kettle, and extracting target poses for robotic grasping.}
\label{fig:overall system}
\end{figure}

\subsection*{Force Feedback}

\begin{figure}[H]
\centering
\includegraphics[width=\linewidth]{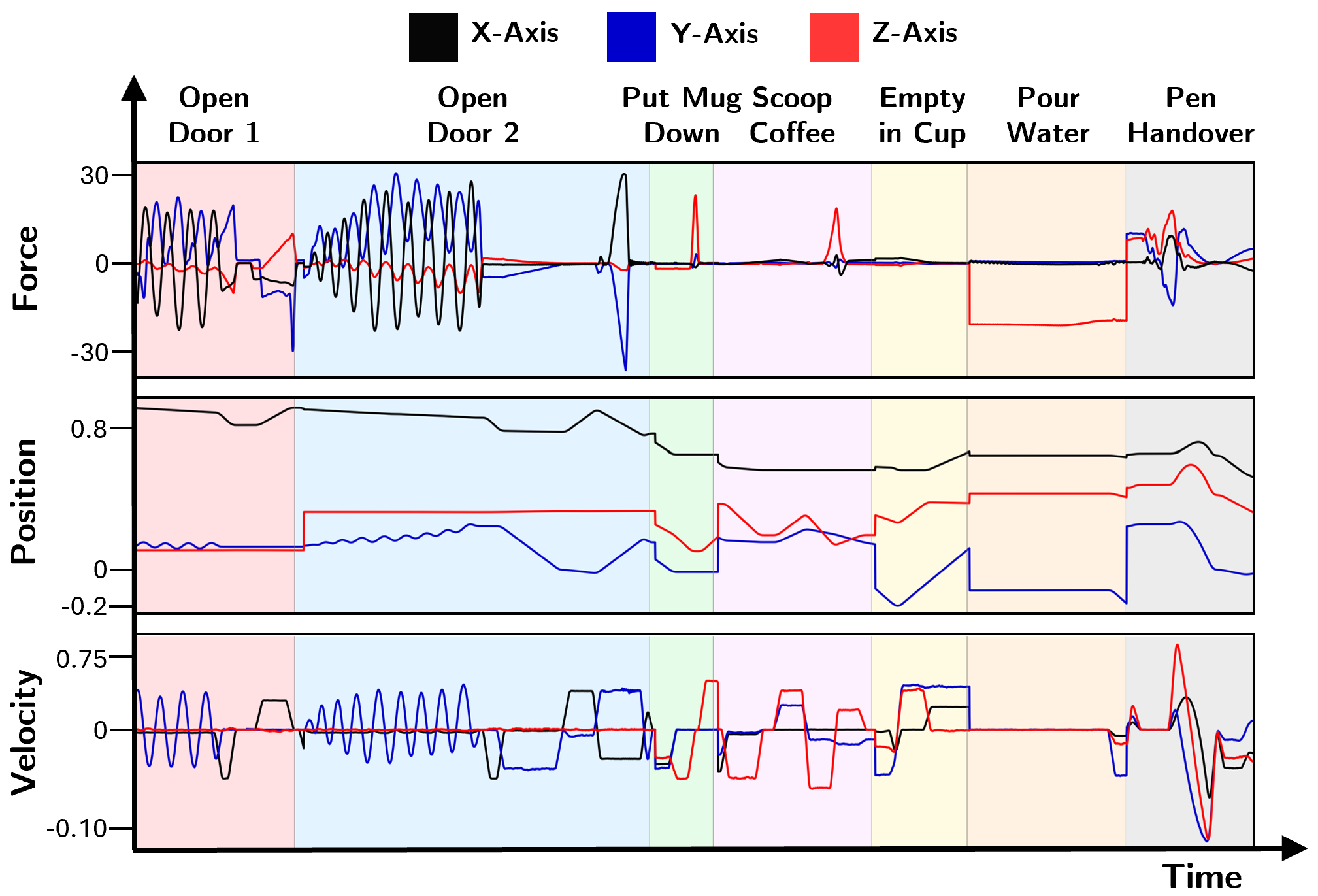}
\caption{Force ($N$), velocity ($\frac{m}{s}$), and position ($m$) plots during a robot's coffee preparation, illustrating diverse force feedback across different motions. Drawing was left out for clarity.}
\label{fig:force plot}
\end{figure}

During task execution, the robot demonstrated a variety of motion dynamics accompanied by distinct types of force feedback. Figure \ref{fig:force plot} illustrates the forces experienced whilst the robot is preparing coffee and handing over a pen. As shown in Figure \ref{fig:force plot}, a diverse spectrum of external forces is handled across various tasks. For example, when putting down a mug, the peak upward force is used as an indicator of successful placement. In contrast, during drawer manipulation, the forces and torques along the x and y axes are critical, highlighting their significance for successful task execution. The variability in force feedback exemplifies the advantages of a scalable approach that adapts to the requirements of diverse motions.

The pouring accuracy achieved was $\sim 5.4$ grams per 100 grams at a pitch velocity of $4 m/s$. We assumed a quasi-static equilibrium to estimate the volume of water poured at any given moment. However, as the pitch velocity increased, the accuracy decreased, with errors approaching $\sim20$ grams per second at a pitch velocity of $30 m/s$. This decrease in accuracy can be attributed to the breakdown of the quasi-static assumption and the impact of the mass distribution of both the pouring medium and container on measurement accuracy.

\subsection*{Generating Art}

\begin{figure}[H]
    \centering
    \begin{subfigure}[t]{\textwidth}
        \centering
        \includegraphics[height=4cm]{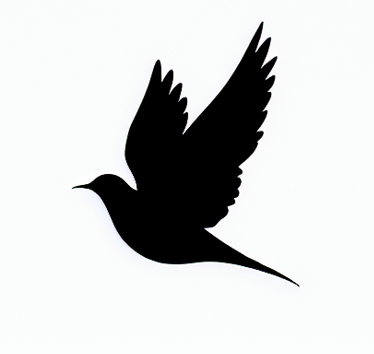} 
        \hfill 
        \includegraphics[height=4cm]{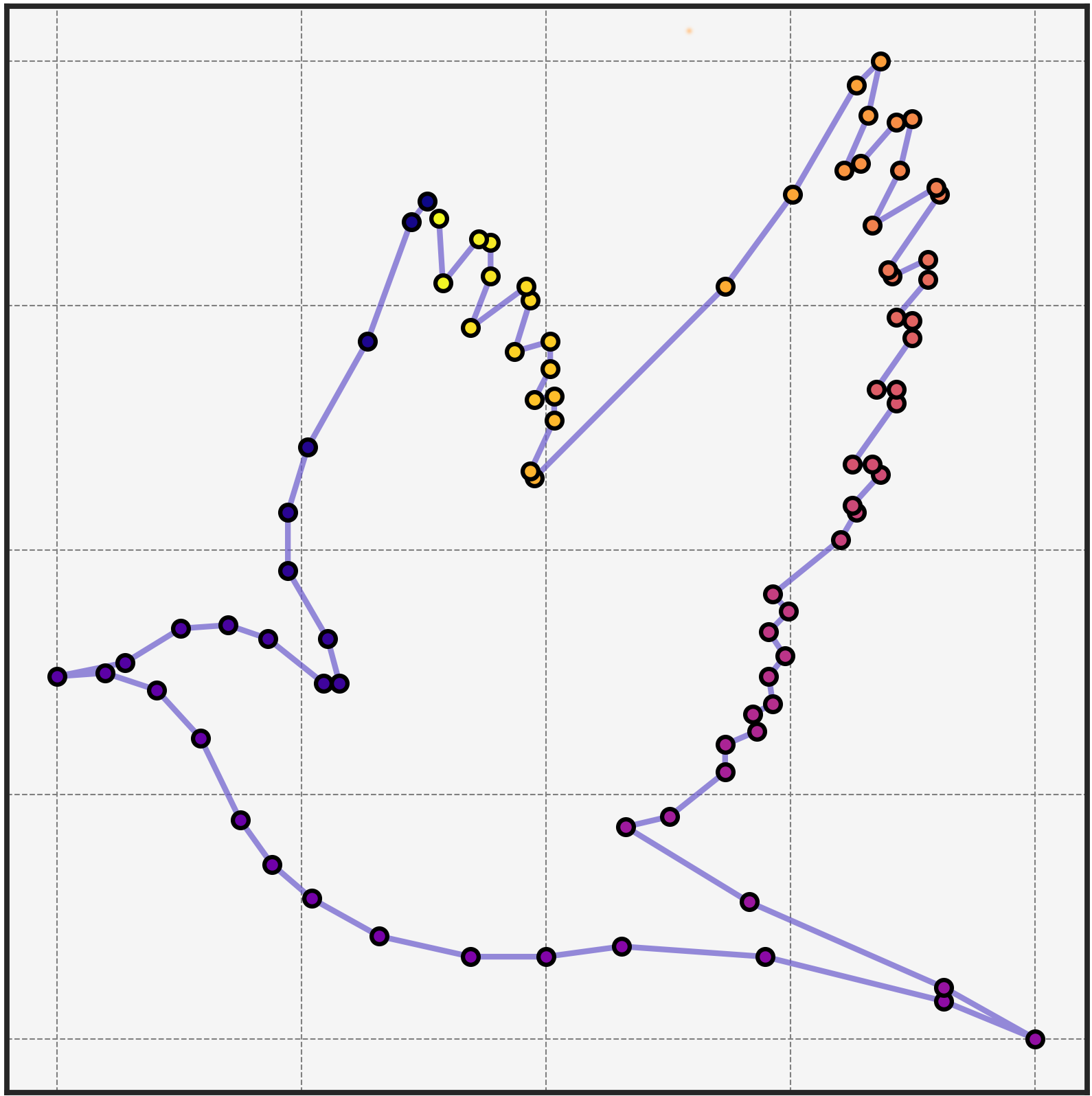} 
        \hfill 
        \includegraphics[height=4cm]{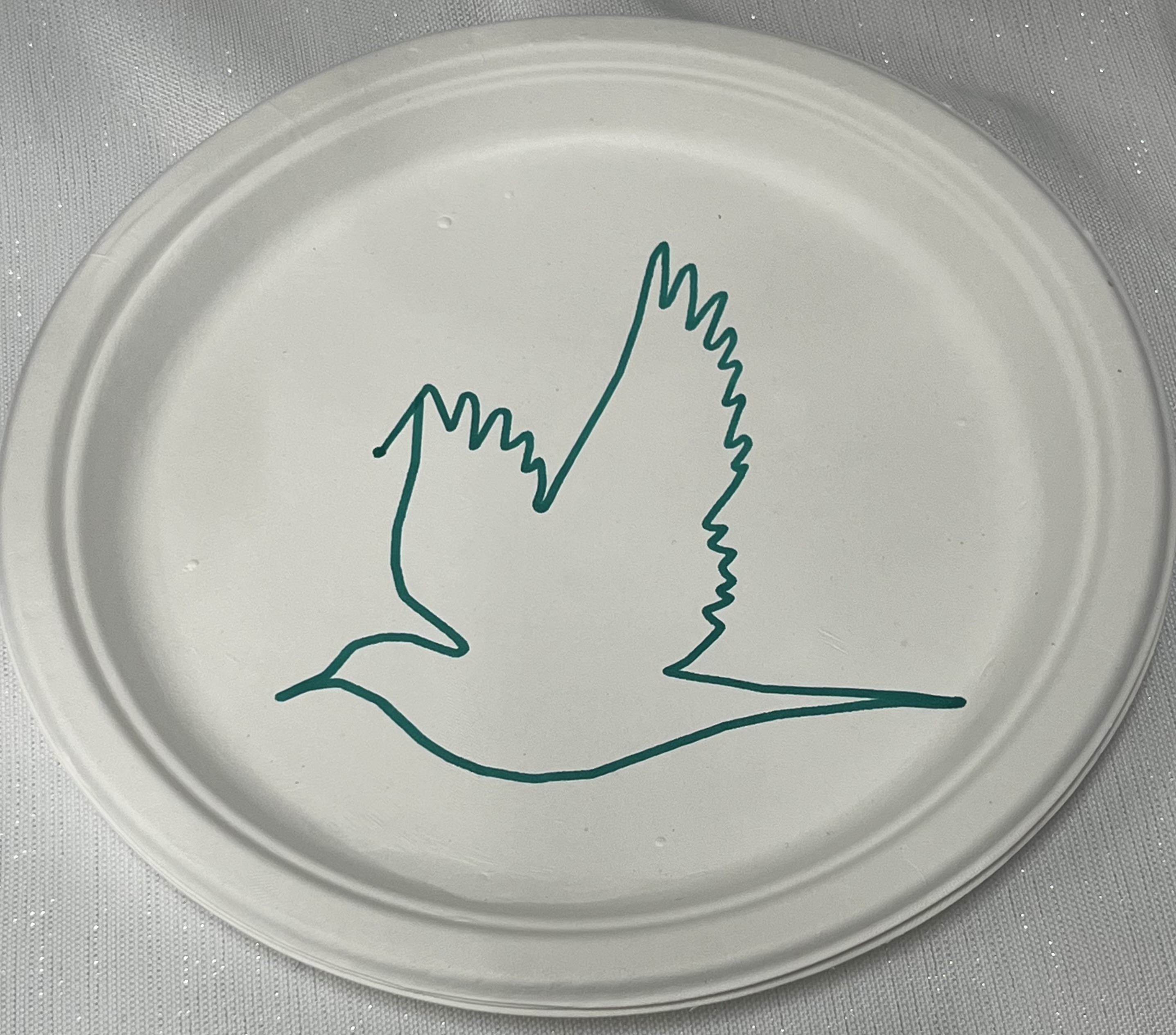} 
        \caption{Generated image, contour plot and drawing of a 'Random Animal'}
        \label{fig:random_animal}
    \end{subfigure}
    
    \vspace{2ex} 
    
    \begin{subfigure}[t]{\textwidth}
        \centering
        \includegraphics[height=4cm]{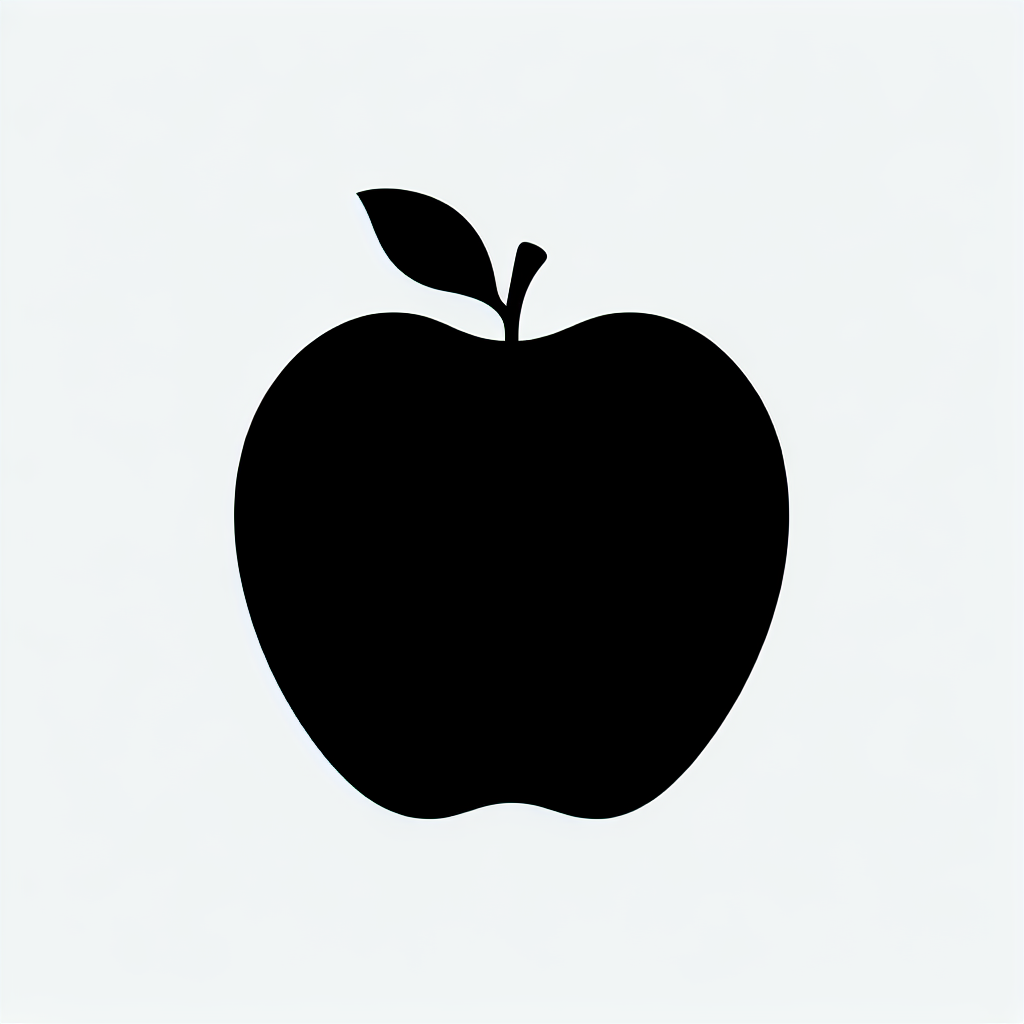} 
        \hfill 
        \includegraphics[height=4cm]{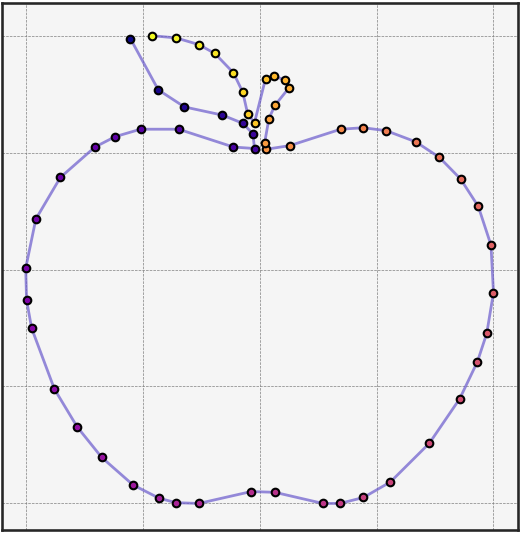} 
        \hfill 
        \includegraphics[height=4cm]{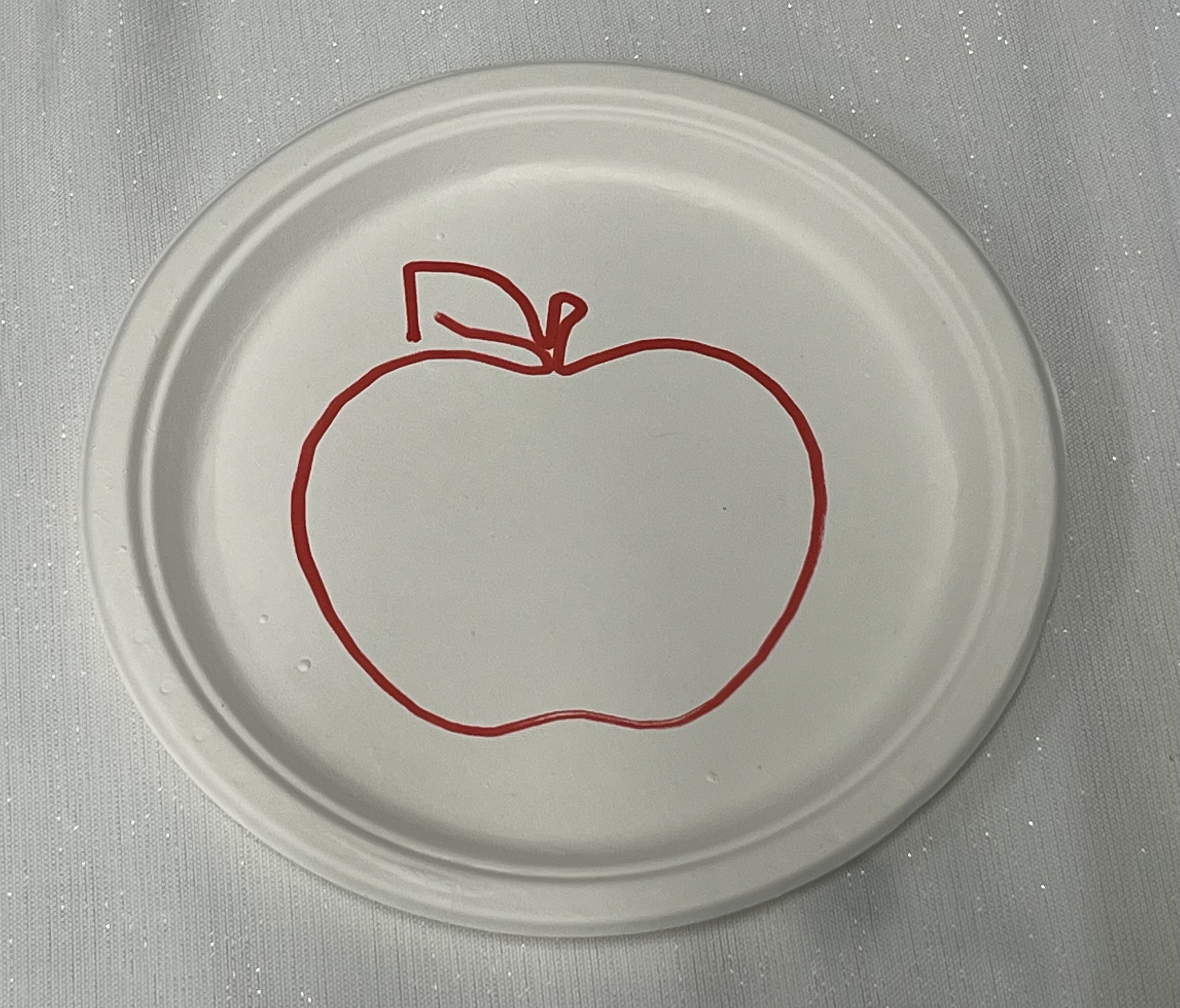} 
        \caption{Generated image, contour plot and drawing of a 'Random Food'}
        \label{fig:random_food}
    \end{subfigure}
    
    \vspace{2ex} 
    
    \begin{subfigure}[t]{\textwidth}
        \centering
        \includegraphics[height=4cm]{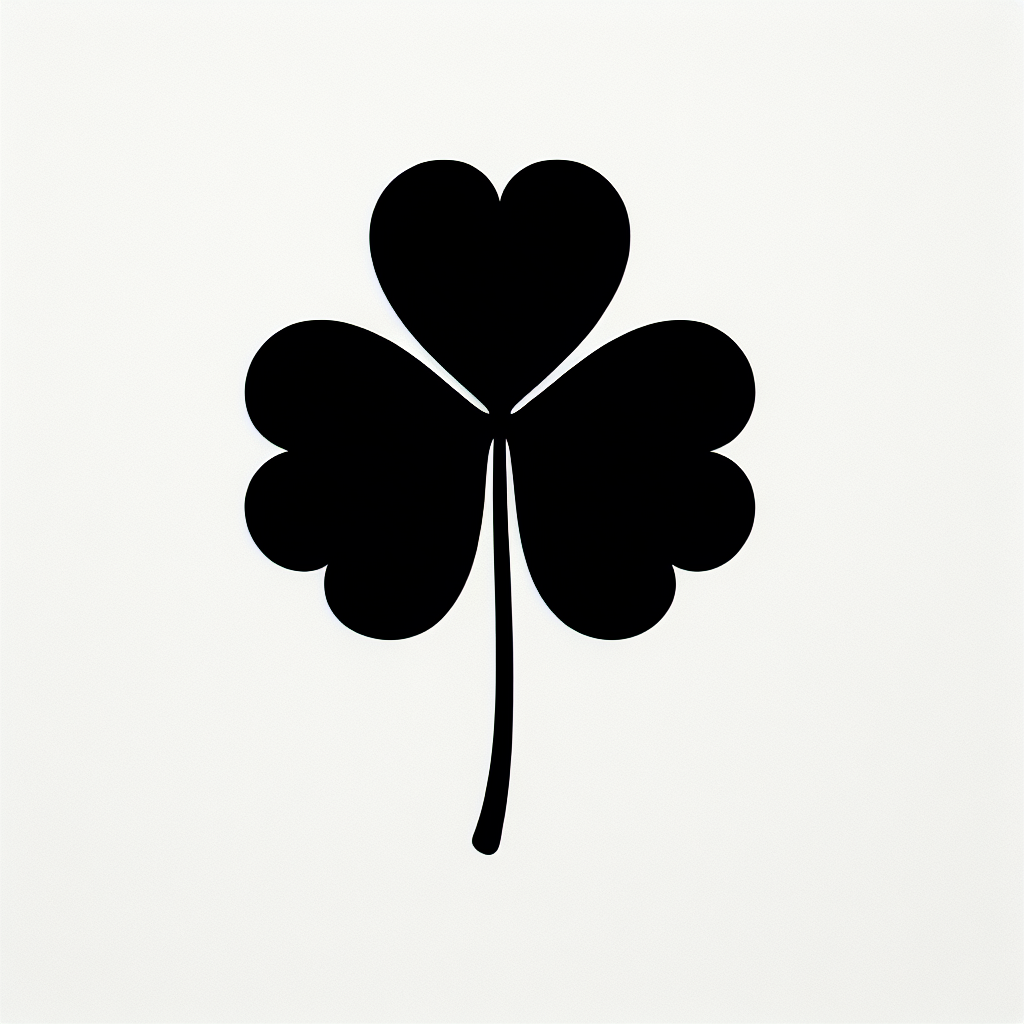} 
        \hfill 
        \includegraphics[height=4cm]{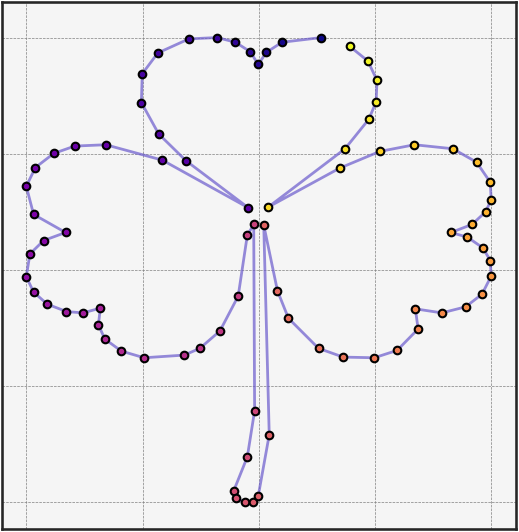} 
        \hfill 
        \includegraphics[height=4cm]{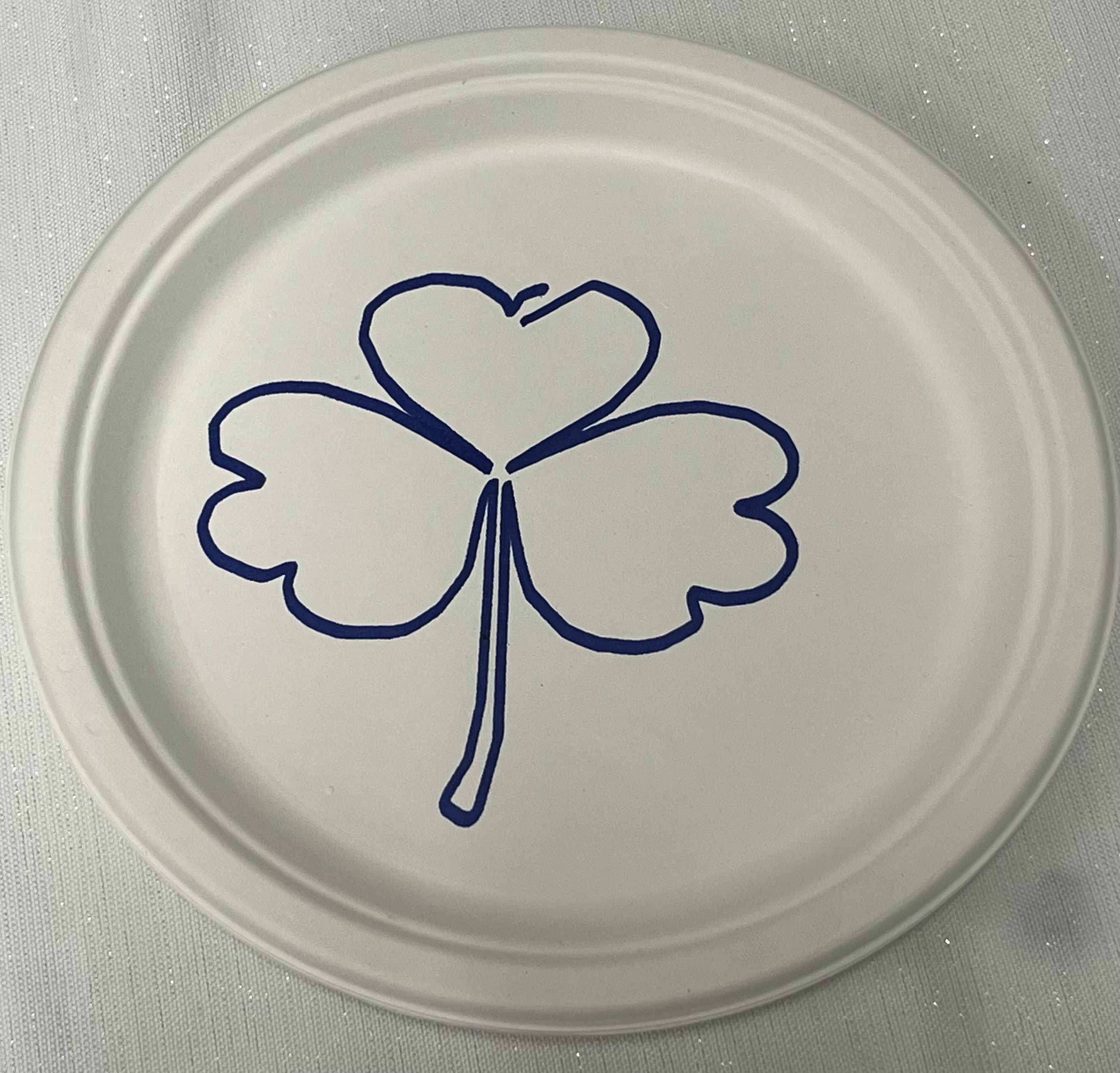} 
        \caption{Generated image, contour plot and drawing of a 'Random Plant'}
        \label{fig:random_plant}
    \end{subfigure}
    
    \caption{Illustration of the drawing process across different queries.}
    \label{fig:drawing_process_all}
\end{figure}

To enable the robot to draw any design specified by the user, we used DALL-E \cite{ramesh_zero-shot_2021} to produce an image from which we could derive a drawing trajectory. This method uses DALL-E to create silhouettes based on keywords extracted from the user, such as “random bird” or “random plant”. The silhouette’s outline is extracted and transformed to match the dimensions of the target surface. This allows the robot to replicate the design on various physical objects as illustrated in Figure \ref{fig:drawing_process_all}. The z-component adjustment is controlled through force feedback to apply an even pen pressure when drawing. Utilising force in the z-axis was important for enabling the pen to apply the desired pressure, even amidst uncertainties in the object’s surface position relative to the camera. In principle, this technique can be adapted for decorating diverse items, such as cakes and coffees (although this would require precise control over the dispensing mechanism). We found that the quality of DALL-E-generated images varied with the specificity of the input prompt and the capabilities of the language-to-vision model. Although the current implementation focuses on silhouette outlines, the principle can be extended to more complex intricate design. A fixed low-velocity of $0.01 m/s$ was used to draw the designs. Figure \ref{fig:drawing_process_all} illustrates the robot’s accuracy at this constant speed. Table \ref{tab:table_drawing} provides details on the time it took the robot to complete each shape and its similarity to the original (measured using the Jaccard Index). Maintaining a low speed helped ensure precise tracing by preventing overshoots as the robot navigated through the waypoints.

\begin{table}[ht]
\centering
\caption{Accuracy of the plotting methods.}
\begin{tabular}{@{} lcr @{}} 
\toprule 
\textbf{Shapes} & \textbf{Time to completion (S)} & \textbf{Completeness ($\%$)} \\ 
\midrule 
Random Animal & 100.09 & 98.56 \\
Random Food & 79.34 & 99.45 \\
Random Plant & 115.65 & 98.03 \\
\bottomrule 
\end{tabular}
\label{tab:table_drawing}
\end{table}

\subsection*{Limitations}

The assumptions of our robotic system framework are twofold: (i) The vision module can accurately identify and classify objects presented within the scene, and (ii) The robot is equipped with a comprehensive affordance map of the utensil in use.

The performance of our system was primarily limited by the capabilities of the vision module. Although the primary study focus was not on object detection, limitations in detection accuracy and response times hindered optimal task performance. In the coffee scenario, under assumption (ii) we endowed our model with prior knowledge of the affordance of the kettle, spoon, and door handles, but we argue that affordances can be learned with very little data from recent work \cite{li_locate_2023, li_one-shot_2024}. We dynamically tracked the pose of the hand and the white mug. Additionally, we presupposed that the robot’s low-level control mechanism can manage obstacle avoidance.

We did not have a quantitative comparison to baseline methods that rely on LLMs because they are primarily designed for short-horizon tasks and lack the variability needed for performing actions ranging from pouring to scooping and drawing  \cite{huang_voxposer_2023}. Moreover, most of these methods lacked IFVF, which makes them unable to deal with several uncertainties in our scenario and would require extensive additional training to handle the uncertainties and differences in our robotic setup \cite{brohan_rt-2_2023}.




\section*{Discussion}
This research investigated the integration of LLMs with robotic systems to enhance their ability to perform complex abstract tasks such as making coffee and decorating plates. By generating specific code, LLMs allow robots to leverage tailored functions, thereby increasing their expressiveness and generalisation capabilities. Motion modules can be easily added from methods such as reinforcement learning and imitation learning, without sacrificing performance.

This work shows how integrating vision, force and language modalities can yield promising results in manipulation tasks. For example, force sensors improve the precision of tasks such as pouring a precise and accurate amount of liquid, while the vision system identifies object positions and movements. The language capabilities enabled the system to produce feedback within the code, which is critical for adjusting to new tasks.

Nonetheless, while LLMs are capable of generating code that addresses known uncertainties to facilitate quick adjustments to disturbances, the adaptations are primarily reactive rather than proactive. The system can use vision and force sensors to adjust to real-time changes, such as scooping more coffee or accessing different draws, but it struggles with proactive adaptations like switching tasks part way through without prior programming. More frequent querying of the language model could allow it to reassess and modify its overall plan based on new inputs. Despite progress, challenges remain. Sophisticated modelling (such as modelling the flow rate as a function of the pouring rate, container size and liquid viscosity) are needed to handle complex force dynamics at the end effector, and spatial awareness tools like OctoMaps (a robotic library for a 3D occupancy map) need to be integrated. Additionally, the accuracy of vision models is essential for real-world application viability.

The framework’s potential is vast, including the method for trajectory generation. For instance, using a model like DALL-E to derive trajectories from visual inputs opens avenues for robotic trajectory generation. The current method can be applied to any object, such as drawing, cake or coffee decoration. For future work, it would be particularly powerful in applications when not only a query but also an image is inputted and can be edited to allow for novel trajectory generation. Moreover, recent enhancements in models like GPT-4O are set to significantly improve the fluidity and effectiveness of interactions.

Overall, the approach demonstrates immense potential for developing efficient, reliable, and highly adaptable robotic systems. Integrating advanced models and control strategies allows robotics to leverage the exponential advancements in LLMs, enabling more sophisticated interactions. This will usher in the next age of automation with unprecedented levels of autonomy and precision, accentuating ¬the need to manage these advancements safely \cite{bengio_managing_2024}.

\section*{Materials and Methods}
\subsection{Overview}
The goal of the robot was to respond to high-level human commands in a dynamic environment, such as a home kitchen. We designed a realistic setting featuring items including a kettle, drawer, and coffee pot. The scenario was designed to demonstrate the robot’s ability to perform diverse tasks in a realistic, albeit reasonably constrained, environment while interacting with a human present. The pipeline consists of a language processing component for task execution, a vision system for pose detection, and a force module for object manipulation. This is all integrated within a Robotic Operating System (ROS) process.

\subsection*{Hardware and Software}
A Kinova 7 degree-of-freedom robot was used. An Azure Kinect Sensor was used at a resolution of 640 $\times$ 576 px and 30 FPS, along with an ATI multi-axis force sensor. A 140mm Robotiq gripper was attached to the end of the robot. The force sensor was attached to the Robotiq gripper and Kinova arm using a 3D printed flange. A small cylinder was placed on the force sensor on the side closest to the gripper so that movements of the gripper would not touch the force sensor, leading to reading being inaccurate. A Dell desktop computer with an Intel Core i9 processor with an NVIDIA RTX 2080-GPU was used and connected to the robot with an Ethernet cable. Similarly both Azure cameras were attached to the desktop. Ubuntu 20.04 and the robotics operating system (ROS) were used. Our code relied on the Kinova ROS Kortex library.



\subsection*{Language Processing}

The LLM processes an image and the user’s query, systematically breaking down the complex task $L_T$ into a sequence of steps $\{L_1, L_2, \ldots, L_N\}$, where each step $L_i$ may depend on the completion of preceding steps. The sequence of steps is critical, and dependencies exist between steps; for example, if an object (e.g. a mug) is required but not found, then potentially a cupboard should be opened.

The environmental data gathered from the initial image input is key in decomposing the abstract task. For instance, when asked to make a beverage, the ingredients present in the environment are critical in deciding which one to make, and the visual information can help identify possible locations. The interface is facilitated by a GPT-4, which runs under the instruction to write and dispatch code to a robot via the server platform “kinovaapi.com’. The process is assisted by a knowledge base containing code examples and allows continuous communication with the robot. The curated knowledge base contains validated examples of low and high-order actions that incorporate known uncertainties. Including these motion examples is key to enabling the robot to handle numerous scenarios and complete long-horizon tasks. High-level motion primitives or policies can compress multiple known uncertainties into a single function, reducing the need for extensive code writing. Retrieval-augmented generation allows the knowledge base to be comprehensive without sacrificing performance. The system interacts with ROS and communicates via a low-latency connection provided by the EC2 server (establishing a reliable link through kinovaapi.com) through JSON action queries and responses.

The dependency among tasks is expressed through conditional probabilities such as $P(L_{2A}, L_{2B} \mid L_1)$, which specifies the likelihood of progressing to tasks $L_{2A}$ or $L_{2B}$ following the successful execution of task $L_1$. This helps in planning the sequence of steps, ensuring the robot can adapt its actions based on real-time feedback. The LLM generates executable code that is sent to the server, based on the instructions (prompt) and a knowledge base containing examples. The code is run on ROS in a secure environment that only has access to predefined functions, thereby ensuring safety in the task execution.

\subsection*{Vision System}
Grounded Segment Anything was used as the language-to-vision model to create a 3D voxel that highlights the positions of all objects and from which their poses can be extracted for robotic grasping \cite{liu_grounding_2023, kirillov_segment_2023}. This enabled (i) the generation of object-specific bounding boxes, (ii) the creation of segmented masks via Mobile Sam and (iii) the creation of voxels that encapsulate detected objects. The voxels allow target object poses to be extracted.

\subsection*{Force Module}
To ensure accurate measurements in force-rich applications, we calibrated the ATI force sensor to compensate for gravitational forces, ensuring it registers zero in the absence of external forces. This calibration is key for accurately predicting the external forces applied to the end effector. The process involved sequentially zeroing the force sensor on one axis, rotating the sensor, and then zeroing the next axis. The local forces were transformed into the global plane to estimate the upward force at different rotations $F_{global} = T_{end\_effector\_to\_robot\_base} \cdot F
_{local}$, where $F_{global}$ is the force vector in the global (robot base) coordinate frame, $T_{end\_effector\_to\_robot\_base}$ is the tranformation matrix from the end effector's frame to the robot's base frame, and $F_{local}$ is the force vector in the local coordinate frame of the end-effector. We explored various methods, such as moving the sensor’s position and orientation and employing polynomial functions for calibration. However, the simpler calibration method was found to be most effective.

To estimate flow rates, we assumed a condition of static equilibrium and maintaining slow operational speeds during pouring. Mathematically, this is represented mathematically as $F_{up} \approx mg$ and $\Delta F_{up} \approx \Delta mg$. In situations involving variable acceleration, the relationship between forces and flow rates becomes more complex. It necessitates a dynamic model that accounts for varying inputs such as the flow rates, the container’s centre of mass, and the inertia of the end effector to map dynamic force inputs to the pouring flow rates.

\subsection*{ROS Operation }

In this work, we initiated the robotic processes by launching a Kinova ROS Kortex driver. This establishes a node that enables communication within the ROS network and the Kinova Gen3 robot. The node publishes several topics that subscribers can access, and it provides services that can be called to modify the robot’s configuration. The base joints are updated at a frequency of 40 Hz. Concurrently, the Robotiq 2F-140mm gripper node is activated at 50 Hz. The node sets up a communication link with the gripper via a USB connection and it initiates an action server that enables precise control of the gripper and facilitates the exchange of operating data. 

A vital element of our robotic system is the vision module node. A ‘classes’ variable is used to identify the target pose of selected objects within the environment. This variable can be dynamically updated, thus allowing the system to adapt to changes in the scene. The pose coordinates of the objects, as established by the ‘classes’ variable, are published approximately at every $\sim \frac{1}{3}$ Hz. This is largely due to the processing time of Grounding DINO in detecting objects and establishing bounding boxes. Moreover, we used an AprilTag to determine the position of the camera relative to the robot’s base. This is represented as $P^R = T_{AR} \cdot (T_{CA} \cdot P^C)$, where $P^C$ is the point in the camera frame, $T_{CA}$ is the transformation matrix from the camera frame to the April tag, $T_{AR}$ is the transformation matrix from the April tag to the robot’s base and PR is the point in the robot’s base frame. 

In parallel, a force node is launched and provides multi-axis force and torque readings at a frequency of 100 Hz, localised to the ATI force transducer. The readings are transformed using a quaternion-based 3x3 rotation matrix to align with the global base frame of the robot, providing raw and averaged values over the last five timesteps across fixed degrees-of-freedom. It calculates forces in the global frame of the robot base using the rotational matrix, calculated from kinematic data.

ROS facilitates the continuous processing of multimodal feedback data from the language processing, vision systems, force metrics and joint end-effector positions. The motions operate on a foundational six degree of freedom twist command, which controls velocity and the variable-speed and force gripper procedures for opening and closing. This enables the integration of hard-coded safety constraints, such as maximum velocity and force limits, as well as workspace boundaries.

The linear velocities are clamped between $\pm 0.05 \si{\meter\per\second}$ and the angular velocities are clamped between  $\pm 60\ \si{\degree\per\second}$. End-effector forces were also limited to \SI{20}{\newton}. This is coded into the fundamental motion primitives so error in the language model will not override this. The end effector is also clamped within the predefined workspace bounds of $x = [0.0, 1.1]$, $y = [-0.3, 0.3]$, $z = [0, 1.0]$. This is checked in future timesteps by a publisher at a frequency of  \SI{10}{\hertz}.

\section*{Acknowledgments}
This work was supported by the EPSRC CDT in RAS (EP/L016834/1). We would like to thank S. Vijayakumar for his support and for providing access to exceptional resources, as well as J. Wang, T. Stouraitis, J. Ferrandis, and many others for their invaluable support and expertise.

\bibliography{ref}

\bibliographystyle{plain}

\clearpage

\end{document}